\documentclass[journal]{IEEEtran}

\usepackage{amsmath,amssymb,amsfonts,bm}
\usepackage{graphicx}
\usepackage{cite}
\usepackage{url}
\usepackage{array}
\usepackage{multirow}
\usepackage{xcolor}
\usepackage{pifont}
\newcommand{\cmark}{\ding{51}}
\newcommand{\xmark}{\ding{55}}

\newcommand{\R}{\mathbb{R}}
\newcommand{\Xset}{\mathcal{X}}
\newcommand{\Acal}{\mathcal{A}}
\newcommand{\PtwoLLM}{P$^2$-LLM}
\newcommand{\softmax}{\operatorname{softmax}}

\begin{document}

\title{LUMI: Tokenizer-Agnostic LLM-Based\\ Lossless Image Compression}

\author{
Chris Xing TIAN,
Chengkai WU,
Ziyu WANG,
Rongqun LIN,
Kecheng CHEN,
Xiandong MENG,\\
Haoliang LI,
Shiqi WANG,
and Siwei MA,~\IEEEmembership{Fellow, IEEE}

\thanks{
Chris Xing TIAN, Rongqun LIN, and Xiandong MENG are with Peng Cheng Laboratory, Shenzhen, China.

Chengkai WU, Kecheng CHEN, and Haoliang LI are with the Department of Electrical Engineering, City University of Hong Kong, Hong Kong SAR.

Ziyu WANG and Shiqi WANG are with the Department of Computer Science, City University of Hong Kong, Hong Kong SAR.

Siwei MA is with the School of Computer Science, Peking University, Beijing, China.

Corresponding author: Siwei MA (e-mail: swma@pku.edu.cn).
}
}

% \markboth{IEEE Transactions on Multimedia}%
% {Tian \MakeLowercase{\textit{et al.}}: LUMI: Tokenizer-Agnostic LLM-Based Lossless Image Compression}

\maketitle

\begin{abstract}
Large language model (LLM)-based lossless image compression methods typically represent pixel data through the native text interface of a pretrained model, converting pixel values into token sequences that the LLM processes through its vocabulary head. This design shows that pretrained language models can provide probability estimates for image coding, but it also couples compression to tokenizer behavior, vocabulary-specific numeric tokens, and model-family-specific adaptation. In this paper, we present LUMI (LLM-based Unified Model-agnostic lossless Image compression), a tokenizer-agnostic framework for lossless RGB image compression with frozen LLM backbones. LUMI replaces pixel-as-text tokenization with a pixel embedding module that maps raw intensity and channel information into the continuous embedding space of the LLM. It further introduces intra-patch position encoding to retain two-dimensional spatial structure after flattening, and uses a 256-way prediction head to produce probabilities over the native pixel alphabet. Only the pixel embedding, position encoding, soft-prefix parameters, and prediction head are trained, while the LLM backbone remains fixed. Experiments on natural, medical, and remote-sensing image benchmarks with LLaMA, Qwen, and Gemma backbones show that LUMI provides a unified interface across tokenizer families, achieves competitive compression rates, and improves cross-domain robustness over tokenizer-based LLM compression baselines. These results formulate LLM-based lossless image compression as pixel-space adaptation of frozen foundation models rather than tokenizer-specific language-symbol modeling.
\end{abstract}

\begin{IEEEkeywords}
Large language models, lossless image compression, entropy modeling, frozen foundation models.
\end{IEEEkeywords}

\section{Introduction}

Large language models (LLMs) are increasingly used as general sequence models beyond conventional text generation. In addition to natural language tasks, recent work has studied their use in coding \cite{chen2021evaluating}, reasoning \cite{wei2022chain}, retrieval \cite{lewis2020retrieval}, multimodal understanding \cite{alayrac2022flamingo}, and data compression \cite{deletang2024language}. From the perspective of lossless compression \cite{cover2006elements,witten1987arithmetic}, this direction is motivated by a direct connection between autoregressive likelihood modeling and entropy coding, since a model that assigns accurate conditional probabilities to the next source symbol can be used by an arithmetic coder to obtain short codelengths.

This connection has motivated recent studies on LLM-based lossless image compression. Early approaches condition language models with visual prompts or residual information \cite{du2025visualprompts,bai2024deep}, while pixel-level methods model RGB images as autoregressive sequences and use an LLM to predict the next pixel value. One example is \PtwoLLM{}, which represents pixel intensities as textual numeric tokens, provides prompt-based pixel priors, and adapts the LLM with parameter-efficient tuning for next-pixel prediction \cite{chen2026large}. These studies indicate that pretrained LLMs can be used as entropy models for image compression.

However, the native interface of an LLM is not designed for exact image-symbol coding. A text tokenizer maps strings to vocabulary tokens, whereas a lossless RGB image codec needs probability distributions over the 256 possible values of each channel. When a pixel value is represented as text, the resulting token sequence depends on the tokenizer. For example, the same string can be encoded as one token by one LLM family but as multiple digit or subword tokens by another. This changes the length of the modeled sequence and the probability events used by arithmetic coding. In addition, vocabulary logits are not naturally aligned with the pixel alphabet, and flattened token streams do not explicitly retain color-channel identity or two-dimensional position.

\begin{figure*}[t]
\centering
\includegraphics[width=\textwidth]{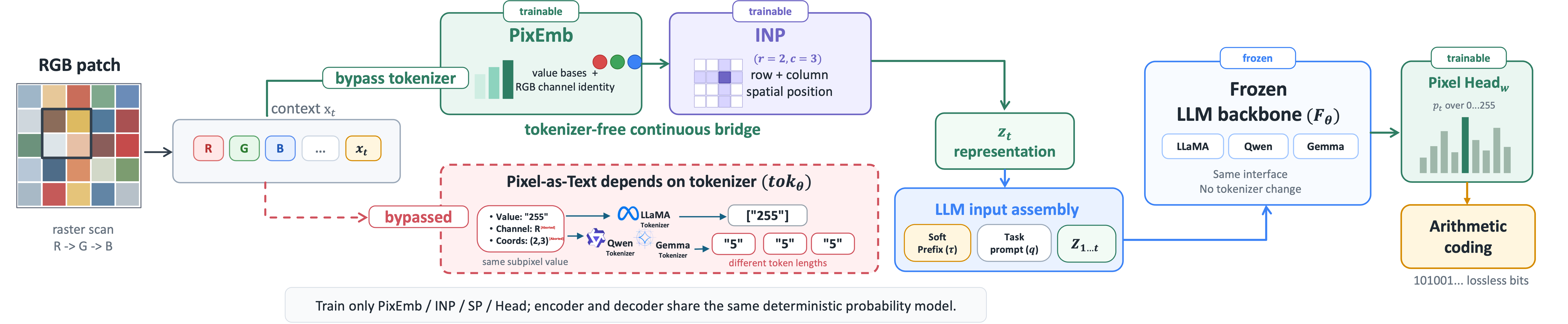}
\caption{Overview of the proposed LUMI framework. A rasterized RGB patch is modeled as a sequence of sub-pixel symbols, while the pixel-as-text route is bypassed to avoid tokenizer-specific fragmentation across LLM families. PixEmb injects intensity and channel information, INP restores intra-patch spatial awareness, and the assembled tokenizer-free representations are processed by a frozen LLM backbone. A lightweight pixel head then predicts a 256-way distribution for arithmetic coding, with only the external interface modules trained.}
\label{fig:overview}
\end{figure*}

These observations motivate a different formulation. Instead of adapting pixel values to a language tokenizer, this paper maps image symbols to the continuous embedding interface of a frozen LLM. We propose \textbf{LUMI} (\textbf{L}LM-based \textbf{U}nified \textbf{M}odel-agnostic lossless \textbf{I}mage compression), a tokenizer-agnostic lossless image compression framework that bypasses textual tokenization for image data. LUMI maps raw pixel intensity, channel identity, and intra-patch spatial position into the LLM embedding space, processes the resulting sequence with a frozen decoder-only backbone, and predicts the next pixel through a 256-way head used by arithmetic coding.
The proposed design keeps the pretrained backbone unchanged and optimizes only a small set of external parameters, including the pixel embedding module, intra-patch position encoding, soft-prefix tokens, and prediction head. This separation matters for applicability across models. The same compression interface can be attached to different LLM families without relying on their numerical-token segmentation or modifying their internal weights.

The contributions of this paper are summarized as follows.
\begin{itemize}
    \item We formulate LLM-based lossless image compression as probability modeling over exact pixel symbols and analyze the tokenizer dependence of existing pixel-as-text formulations.
    \item We propose a tokenizer-free pixel interface that maps raw intensity and channel information into the continuous latent space of frozen LLM backbones and predicts a 256-way pixel distribution for entropy coding.
    \item We introduce intra-patch position encoding to supply local spatial coordinates after flattening, and study its effect on the suitability of frozen LLMs for image-symbol modeling.
    \item We evaluate the method on LLaMA, Qwen, and Gemma backbones over natural, medical, and remote-sensing images, and report the compression rate, cross-domain behavior, and behavior across tokenizer families.
\end{itemize}

\section{Related Work}

\subsection{Lossless Image Compression}

Lossless image compression often combines source modeling with entropy coding. Classical codecs such as PNG \cite{boutell1997png}, JPEG-LS \cite{weinberger2000loco}, and JPEG-XL \cite{alakuijala2019jpeg} reduce redundancy through predictors, reversible transforms or filters, context models, and entropy coders. These methods remain practical baselines because their design choices are closely matched to image statistics.

Learned lossless compression replaces part of this pipeline with neural probability models. Autoregressive density models such as PixelRNN and PixelCNN estimate pixel distributions from causal context \cite{van2016pixel,salimans2017pixelcnn}. Later learned codecs improve compression through hierarchical probability models, lossy-plus-residual coding, invertible flows, and bit-plane modeling \cite{mentzer2019pract, mentzer2020learning, zhang2021ivpf, zhang2021iflow, bai2024deep, zhang2024learned}. These methods show that improved entropy models can reduce codelength, but they are usually trained as image-specialized codecs. LUMI instead studies whether lossless image coding can be implemented as a lightweight interface on top of frozen foundation models.

These methods demonstrate that improved probability modeling is central to reducing lossless codelength. However, most learned codecs are designed as image-specialized systems, with architectures, auxiliary variables, residual models, or bit-level structures tailored to visual data and trained directly for image compression. LUMI follows the same entropy-coding principle but studies a different question. Instead of designing a full image-specific codec, it investigates whether a frozen foundation model can be reused as a contextual entropy model through a lightweight pixel-space interface. In this sense, LUMI shifts the focus from building a specialized neural image codec to adapting exact image symbols to the latent computation space of a general-purpose LLM.

\subsection{Language Models as Compressors}

Autoregressive language modeling is closely related to lossless compression. Under arithmetic coding, the negative log-likelihood assigned to each next symbol determines its ideal codelength. Deletang \textit{et al.} \cite{deletang2024language} demonstrated that language models can act as general-purpose compressors across modalities. Recent studies further investigate compression with pretrained transformers and large generative models on byte-level or multimodal data \cite{heurtel2024compression,li2025lossless}. These results motivate using foundation models as reusable probability models rather than training a separate compressor for every data type.

For image compression, however, a pretrained LLM does not directly provide a complete codec. Its native input is a token sequence produced by a text tokenizer, and its native output is a vocabulary-level distribution. Lossless image coding requires exact probabilities over image symbols. Therefore, an interface is needed to map image data into the LLM input space and map model outputs back to the entropy-coding alphabet.

\subsection{LLM-Based Lossless Image Compression}

Existing LLM-based image compressors mainly bridge the image-text gap through discrete tokenization or task-specific adaptation. Visual-prompt methods use visual representations or lossy reconstructions to condition an LLM and then encode residual information \cite{du2025visualprompts}. Pixel-level methods avoid patch-level loss by treating RGB values as an autoregressive sequence. \PtwoLLM{} represents pixel intensities as textual numeric tokens, uses prompts and pixel priors, and applies LoRA adaptation for next-pixel prediction \cite{chen2026large,hu2021lora}. These methods show that LLMs can provide useful probability estimates for RGB image coding.

Their interface remains tied to language-token design. Tokenizer choices can affect LLM behavior \cite{ali2023tokenizer}. In image compression, this issue becomes concrete because the same pixel value can correspond to different numbers of tokens across model families. Moreover, vocabulary logits are not naturally aligned with the 256-symbol pixel alphabet, and flattened pixel streams discard explicit two-dimensional coordinates. LUMI addresses these limitations by bypassing textual tokenization, embedding pixel values directly in continuous LLM space, injecting spatial coordinates, and predicting probabilities with a dedicated pixel head.

\subsection{Lightweight Adaptation of Frozen Foundation Models}

Parameter-efficient adaptation methods reuse pretrained models while updating only a small number of task-specific parameters. Representative examples include prefix tuning \cite{li2021prefix}, prompt tuning \cite{lester2021power}, and low-rank adaptation \cite{hu2021lora}. These methods reduce adaptation cost and support the use of foundation models as shared computational backbones.

For lossless image compression, the distinction between external adaptation and backbone finetuning is important. Full finetuning or LoRA can improve task performance, but it also specializes the model toward a tokenizer, symbolization scheme, and compression task. LUMI keeps the LLM backbone frozen and trains only external pixel-space modules, a soft prefix, and a probability head. This does not imply that frozen-backbone approaches are expected to exceed all specialized codecs. Rather, it studies how far a shared LLM can be used through a portable tokenizer-free interface.

\section{Methodology}

\subsection{Problem Formulation and Overview}

From the viewpoint of entropy coding, lossless image compression reduces to estimating a conditional probability distribution for the next source symbol. Let an RGB patch be
\begin{equation}
    \mathbf{I} \in \{0,\ldots,255\}^{H_p \times W_p \times 3}.
    \label{eq:image_patch}
\end{equation}

We flatten the patch in row-major order with channel order $R \rightarrow G \rightarrow B$, obtaining
\begin{equation}
    \mathbf{x}=(x_1,x_2,\ldots,x_T), \quad
    T=3H_pW_p, \quad x_t \in \Xset,
    \label{eq:sequence}
\end{equation}
where $\Xset=\{0,\ldots,255\}$ is the 256-symbol source alphabet, i.e., the set of values a single channel can take. The likelihood factorizes as
\begin{equation}
    p(\mathbf{x})=\prod_{t=1}^{T} p(x_t \mid \mathbf{x}_{<t}),
    \label{eq:factorization}
\end{equation}
and the ideal arithmetic-coding length in bits is
\begin{equation}
    L_{\mathrm{bits}}(\mathbf{x})
    =-\sum_{t=1}^{T}\log_2 p(x_t \mid \mathbf{x}_{<t}).
    \label{eq:codelength}
\end{equation}
Thus, compression performance is determined by the quality of the autoregressive entropy model over the 256-symbol pixel alphabet.

\PtwoLLM{} realizes Eq. \eqref{eq:factorization} by converting each pixel value into text and querying a language-model vocabulary distribution \cite{chen2026large}. Following \PtwoLLM{}, the channel-ordered sequence can also be written with an explicit RGB factorization. For a spatial location $(r,s)$, let $\mathcal{C}_{r,s}$ denote all previously decoded spatial and channel context before the red value at that location. Then
\begin{align}
    p(R_{r,s},G_{r,s},B_{r,s}\mid \mathcal{C}_{r,s})
    &=p(R_{r,s}\mid \mathcal{C}_{r,s}) \nonumber\\
    &\quad \cdot p(G_{r,s}\mid \mathcal{C}_{r,s},R_{r,s}) \nonumber\\
    &\quad \cdot p(B_{r,s}\mid \mathcal{C}_{r,s},R_{r,s},G_{r,s}).
    \label{eq:p2llm_rgb_factorization}
\end{align}
This prior factorization is compatible with arithmetic coding and lets the model use inter-channel dependencies within each pixel. However, in \PtwoLLM{} the resulting pixel values are still represented through tokenizer-dependent text-token events. Let $\operatorname{str}(v)$ be the decimal string of a pixel value $v$, and let
\begin{equation}
    \operatorname{tok}_{\theta}(\operatorname{str}(v))
    = (y_{v,1},\ldots,y_{v,m_{\theta}(v)})
    \label{eq:p2llm_tokenization},\quad v \in \Xset.
\end{equation}
be its tokenizer-dependent token sequence for an LLM family with parameters $\theta$, where $m_{\theta}(v)\in\mathbb{N}$ is the number of tokens into which the tokenizer splits the value $v$. Note that $m_{\theta}(v)$ is family-dependent: the same value can yield one token for one tokenizer and several for another, depending on the tokenizer's vocabulary and segmentation rules. A tokenizer-induced pixel probability can then be defined as
\begin{equation}
    p_{\theta}(x_t=v\mid \mathbf{x}_{<t},q)
    \triangleq
    \prod_{j=1}^{m_{\theta}(v)}
    p_{\theta}(y_{v,j}\mid q,\operatorname{tok}_{\theta}(\mathbf{x}_{<t}),y_{v,<j}),
    \label{eq:p2llm_probability}
\end{equation}
where $q$ is the natural-language task prompt introduced in \PtwoLLM{}, specifying the pixel serialization protocol and autoregressive prediction setting for image compression and the language-model distribution $p_{\theta}$ may be obtained from the original backbone or from a parameter-efficiently adapted model.

This formulation is valid as an autoregressive coding model, but it exposes several limitations. First, $m_{\theta}(v)$ depends on the tokenizer, so the same sub-pixel value can become one event in one LLM family and several events in another. Second, vocabulary probabilities are derived from language logits rather than being a native categorical distribution over the 256-symbol source alphabet $\Xset$, which makes pixel coding depend on token selection and normalization details. Third, textual numeric tokens do not explicitly encode channel identity or two-dimensional location. Finally, LoRA or full finetuning can improve the \PtwoLLM{} pathway in terms of likelihood modeling. However, they introduce task-specific parameter coupling, which reduces the reusability of the backbone as a general-purpose entropy model and further entangles the model with a tokenizer-dependent compression interface.

LUMI keeps the probabilistic objective in Eq.~\eqref{eq:codelength} but changes the representation interface. Instead of representing pixel values as strings, it maps pixel symbols directly into the continuous embedding space of a frozen LLM and predicts probabilities through a 256-way pixel head.

Throughout this section, $[\cdot\,;\,\cdot]$ denotes vertical (row-wise) stacking applied to vector or matrix arguments. All components share a common LLM embedding dimension $d$. As illustrated in Fig.~\ref{fig:overview}, for a single patch, LUMI first constructs
\begin{equation}
    \mathbf{E}_{\mathrm{aug}}
    =
    [\bm{\tau};\mathbf{E}_{\mathrm{text}}(q);\mathbf{Z}_{<T}]
    \in\mathbb{R}^{(P+L_q+T-1)\times d},
    \label{eq:lumi_aug_overview}
\end{equation}
where $\mathbf{Z}_{<T}=[\mathbf{z}_1;\ldots;\mathbf{z}_{T-1}]$. The frozen decoder-only LLM produces
\begin{equation}
    \mathbf{H}=F_{\theta}(\mathbf{E}_{\mathrm{aug}}),
    \qquad
    \mathbf{h}_t=\mathbf{H}_{P+L_q+t-1},\quad 1\leq t\leq T,
    \label{eq:lumi_state_selection}
\end{equation}
assuming $P+L_q>0$ so that the final conditioning position predicts the first source symbol. The pixel head then gives
\begin{equation}
    \bm{\pi}_t
    =
    \softmax\!\left(\operatorname{Head}_{\omega}(\mathbf{h}_t)\right)
    \in\mathbb{R}^{256},
    \quad 1\leq t\leq T.
    \label{eq:lumi_overview}
\end{equation}
The distribution $\bm{\pi}_t$ is a categorical distribution over $\Xset$. 
% Because $\mathbf{h}_t$ is computed from the soft prefix, the prompt, and $\mathbf{Z}_{<t}=[\mathbf{z}_1;\ldots;\mathbf{z}_{t-1}]$, the model predicts $x_t$ without observing $x_t$ itself.
The model is trained in an autoregressive manner such that each prediction $\bm{\pi}_t$ depends only on the prefix tokens, the prompt, and the preceding pixel representations, i.e.,
\[
\mathbf{Z}_{<t} = (\mathbf{z}_1,\ldots,\mathbf{z}_{t-1}),
\]
thereby preserving causality in the embedding space rather than the pixel space.
\subsection{Tokenizer-Free Pixel Interface}
\label{sec:pixemb}

The central innovation of LUMI is to replace the tokenizer-bound language interface with a pixel-latent interface on both sides of the frozen LLM. On the input side, pixel values are represented as numerical source symbols with channel identity, not as decimal strings. On the output side, the model predicts the exact entropy-coding alphabet, not selected vocabulary logits. This design makes the LLM a contextual computation module between two pixel-space adapters, allowing the same compression pipeline to be used with different tokenizer families.

For the $t$-th sub-pixel, let $\kappa_t\in\{0,1,2\}$ denote its channel index and let $(r_t,s_t)$ denote its row and column coordinates inside the patch. We normalize the intensity as $u_t={x_t}/{255}\in[0,1]$. Let $\mathbf{e}_{\kappa_t}\in\R^3$ be the one-hot channel indicator. The intensity-channel descriptor is
\begin{equation}
    \mathbf{a}_t=
    \left[
    u_t-\frac{1}{2};
    \left(u_t-\frac{1}{2}\right)^2;
    \sin(2\pi u_t);
    \cos(2\pi u_t);
    \mathbf{e}_{\kappa_t}
    \right]\in\R^7.
    \label{eq:pixel_descriptor}
\end{equation}

This descriptor is designed so that each component captures a complementary aspect of the source symbol. The polynomial terms $u_t-\tfrac{1}{2}$ and $(u_t-\tfrac{1}{2})^2$ provide low-order nonlinear functions of the normalized intensity, giving the MLP direct access to centered magnitude and contrast-related variation. The sinusoidal terms $\sin(2\pi u_t)$ and $\cos(2\pi u_t)$ act as bounded Fourier-style nonlinear features \cite{tancik2020fourier} of the scalar intensity. They enrich the descriptor with smooth non-polynomial responses while keeping the representation deterministic and numerically stable. Finally, the one-hot indicator $\mathbf{e}_{\kappa_t}$ injects explicit color-channel identity, allowing the model to learn channel-specific statistics, such as R/G/B bias and inter-channel correlation, rather than treating all channels identically. Combining these polynomial, sinusoidal, and channel-identity features into a single 7-D vector gives PixEmb a structured numerical input that is easier to project into the LLM latent space than a raw scalar, while remaining independent of any text tokenizer.

The pixel embedding module maps the descriptor into the $d$-dimensional LLM embedding space:
\begin{align}
    \operatorname{PixEmb}_{\phi}&:\R^7\rightarrow\R^d,
    \label{eq:pixemb_map}\\
    \mathbf{e}^{\mathrm{pix}}_t
    &=\operatorname{PixEmb}_{\phi}(\mathbf{a}_t)\in\R^d,
    \label{eq:pixemb}
\end{align}
where $\phi$ denotes the parameters of $\operatorname{PixEmb}_{\phi}$, implemented as a two-layer GELU MLP ($\mathbb{R}^7 \rightarrow 2d \rightarrow d$). This module acts as a lightweight modality adapter that projects structured pixel descriptors into the frozen LLM embedding space, enabling tokenizer-free pixel representation without modifying the tokenizer or backbone.
                        
\subsection{Intra-Patch Position Encoding}
\label{sec:inp}

The second interface requirement is spatial awareness. Flattening creates a valid causal sequence, but it hides the two-dimensional coordinates that explain local image predictability. LUMI augments each pixel embedding with intra-patch position encoding (INP). Since channel identity is already included in Eq. \eqref{eq:pixel_descriptor}, INP focuses on spatial location.

To restore two-dimensional coordinates after flattening, INP assigns one trainable embedding to each distinct row index and one to each distinct column index of the patch, collected in a row-position table $\mathbf{R}\in\R^{H_p\times d}$ and a column-position table $\mathbf{C}\in\R^{W_p\times d}$. The $r$-th row $\mathbf{R}_{r}\in\R^d$ is the embedding of horizontal coordinate $r\in\{1,\ldots,H_p\}$, and the $s$-th row $\mathbf{C}_{s}\in\R^d$ is the embedding of vertical coordinate $s\in\{1,\ldots,W_p\}$. The two tables are learned so that each row index and each column index acquires a $d$-dimensional spatial code shared by all sub-pixels on that row or column. For a sub-pixel at $(r_t,s_t)$, its spatial vector is the gated sum of the corresponding row and column codes,
\begin{equation}
    \mathbf{s}^{\mathrm{pos}}_t
    =\beta(\mathbf{R}_{r_t}+\mathbf{C}_{s_t})\in\R^d,
    \label{eq:pos_encoding}
\end{equation}
where $\beta\in\R$ is a learnable scalar gate that controls the overall strength of the position signal. The additive form reflects the assumption that the spatial code of a position factorizes into an independent row contribution and an independent column contribution, which is the standard bias behind row-column (or axial) position encoding. The final pixel representation combines the intensity-channel content with this spatial code:
\begin{equation}
    \mathbf{z}_t=\mathbf{e}^{\mathrm{pix}}_t+\mathbf{s}^{\mathrm{pos}}_t\in\R^d.
    \label{eq:z}
\end{equation}
Stacking over the patch, the per-pixel representation matrix is $\mathbf{Z}=[\mathbf{z}_1;\ldots;\mathbf{z}_{T}]\in\R^{T\times d}$, with the shifted prefix $\mathbf{Z}_{<T}=[\mathbf{z}_1;\ldots;\mathbf{z}_{T-1}]\in\R^{(T-1)\times d}$ used as the causal input to the LLM (Eq.~\eqref{eq:lumi_overview}). Compared with a full $H_pW_p$ positional table, the row-column decomposition in \eqref{eq:pos_encoding} uses only $(H_p+W_p)d+1$ parameters and remains deterministic for both arithmetic encoding and decoding.

\subsection{Frozen LLM Entropy Model}
\label{sec:frozen}

With the tokenizer-free visual interface, the frozen LLM is used as a contextual entropy model. LUMI prepend a fixed textual task prompt $q$, embedded by the native text embedding layer as
\begin{equation}
    \mathbf{E}_{\mathrm{text}}(q)
    =[\mathbf{e}^{\mathrm{text}}_1;\ldots;\mathbf{e}^{\mathrm{text}}_{L_q}]
    \in\R^{L_q\times d},
    \label{eq:text_prompt}
\end{equation}
where each $\mathbf{e}^{\mathrm{text}}_i\in\R^d$. It also prepends a trainable soft prefix $\bm{\tau}\in\R^{P\times d}$, which is shared by all patches and acts as a compact task conditioner.

To preserve causality, the LLM receives the shifted pixel sequence
\begin{equation}
    \mathbf{E}_{\mathrm{aug}}
    =[\bm{\tau};\mathbf{E}_{\mathrm{text}}(q);\mathbf{Z}_{<T}]
    \in\R^{(P+L_q+T-1)\times d}.
    \label{eq:augmented_input}
\end{equation}
Let $F_{\theta}$ be the decoder-only LLM with frozen parameters $\theta$. The contextual hidden states are
\begin{equation}
    \mathbf{H}=F_{\theta}(\mathbf{E}_{\mathrm{aug}})
    \in\R^{(P+L_q+T-1)\times d}.
    \label{eq:hidden_states}
\end{equation}
Using one-based indexing and writing $\mathbf{H}_i$ for the $i$-th row of $\mathbf{H}$, the hidden state for predicting the $t$-th source symbol is the row $\mathbf{h}_t=\mathbf{H}_{P+L_q+t-1}\in\R^d$ for $1\leq t\leq T$. When no textual prompt is used, $L_q=0$ and the first symbol is predicted from the final soft-prefix position.

The output side is detached from the language vocabulary through a dedicated pixel head:
\begin{equation}
    \bm{\pi}_t = \operatorname{softmax}\!\big(\operatorname{Head}_{\omega}(\mathbf{h}_t)\big), 
    \quad \operatorname{Head}_{\omega}: \mathbb{R}^d \to \mathbb{R}^{256}.
    \label{eq:head_softmax}
\end{equation}

where $\omega$ collects the parameters of $\operatorname{Head}_{\omega}$ (a two-layer GELU MLP from $\R^d$ to $\R^{256}$). The conditional probability used by arithmetic coding is
\begin{equation}
    p_{\Acal}(x_t=v\mid \mathbf{x}_{<t},q)=\bm{\pi}_t[v],
    \quad v\in\Xset,
    \label{eq:conditional_probability}
\end{equation}
where $\Acal$ denotes all trainable adaptation parameters. Together, $\operatorname{PixEmb}_{\phi}$ and $\operatorname{Head}_{\omega}$ form an input-output alphabet alignment layer: pixels enter and leave the model in source-symbol space, while the frozen LLM supplies contextual computation.

\subsection{Training Objective}
 
Only the external adaptation parameters are optimized, namely $\Acal=\{\phi,\psi,\omega,\bm{\tau}\}$ with $\psi=\{\mathbf{R},\mathbf{C},\beta\}$: the PixEmb parameters $\phi$, the INP parameters $\psi$, the pixel-head parameters $\omega$, and the soft prefix $\bm{\tau}$. The frozen LLM parameters $\theta$ are not updated. For a batch of $B$ patches, teacher forcing with shifted inputs gives the negative log-likelihood
\begin{equation}
    \mathcal{L}_{\mathrm{train}}(\Acal)
    =-\frac{1}{BT}\sum_{b=1}^{B}\sum_{t=1}^{T}
    \log p_{\Acal}(x_t^{(b)}\mid \mathbf{x}_{<t}^{(b)},q).
    \label{eq:train_loss}
\end{equation}
This is standard cross-entropy over 256 source symbols, measured in nats when $\log$ denotes the natural logarithm. The optimization problem is $\min_{\phi,\psi,\omega,\bm{\tau}}\mathcal{L}_{\mathrm{train}}(\Acal)$. Expressed in bits, the same objective estimates the expected symbol codelength:
\begin{equation}
    \mathcal{L}_{\mathrm{bits}}(\Acal)
    =-\frac{1}{BT}\sum_{b=1}^{B}\sum_{t=1}^{T}
    \log_2 p_{\Acal}(x_t^{(b)}\mid \mathbf{x}_{<t}^{(b)},q).
    \label{eq:bits_loss}
\end{equation}

\subsection{Arithmetic Coding and Lossless Decoding}

At inference time, LUMI supplies the probability model for arithmetic coding. For each patch, the encoder processes symbols sequentially. At step $t$, it constructs $\mathbf{x}_{<t}$, computes the distribution $\bm{\pi}_t=p_{\Acal}(\cdot\mid \mathbf{x}_{<t},q)$, and encodes $x_t$ under $\bm{\pi}_t$. The ideal patch codelength is
\begin{equation}
    L_{\mathrm{bits}}(\mathbf{x})
    =-\sum_{t=1}^{T}\log_2 \bm{\pi}_t[x_t].
    \label{eq:patch_bits}
\end{equation}
The corresponding bits per pixel (BPP) for an RGB patch is
\begin{equation}
    \operatorname{BPP}(\mathbf{x})
    =-\frac{1}{H_pW_p}\sum_{t=1}^{T}\log_2 \bm{\pi}_t[x_t]
    =-\frac{3}{T}\sum_{t=1}^{T}\log_2 \bm{\pi}_t[x_t].
    \label{eq:bpp}
\end{equation}

Decoding mirrors the same procedure. At step $t$, the decoder has recovered $\hat{\mathbf{x}}_{<t}=\mathbf{x}_{<t}$ and therefore reconstructs the same pixel embeddings, position encodings, soft prefix, and prompt. Because all of these depend only on the shared trainable parameters $\Acal=\{\phi,\psi,\omega,\bm{\tau}\}$ (PixEmb $\phi$, INP $\psi=\{\mathbf{R},\mathbf{C},\beta\}$, the pixel head $\omega$, and the soft prefix $\bm{\tau}$) while the frozen LLM parameters $\theta$ are identical on both sides by construction, the decoder evaluates exactly the same distribution as the encoder, i.e., $p_{\Acal}(\cdot\mid \hat{\mathbf{x}}_{<t},q)=p_{\Acal}(\cdot\mid \mathbf{x}_{<t},q)$. The arithmetic decoder can thus recover $x_t$ from the bitstream under the identical probability model. Repeating this process exactly reconstructs the original patch sequence. The lossless property does not require $\operatorname{PixEmb}_{\phi}$ or the LLM hidden mapping to be invertible. It only requires the encoder and decoder to share the same deterministic probability model and the same symbol order.

\subsection{Discussion of the Adaptation Design}

LUMI differs from tokenizer-dependent LLM compressors in both representation and adaptation. On the input side, it writes pixels directly into continuous embedding space rather than translating them into textual numbers. On the output side, it predicts the exact 256-symbol source alphabet rather than extracting probabilities from vocabulary tokens. These two changes remove the need for tokenizer-specific event definitions and make the same interface applicable to heterogeneous LLM families.

The frozen-backbone design is also relevant. Rather than specializing the LLM through full finetuning or LoRA, LUMI attaches a small number of trainable interface parameters around a shared foundation model. This formulates lossless image compression as a lightweight capability of an existing sequence model. The current implementation compresses non-overlapping patches independently for manageable context length and parallelism. Exploiting inter-patch context is left for future work.

\section{Experiments}
\begin{figure*}[htbp]
\centering
\includegraphics[scale=0.47]{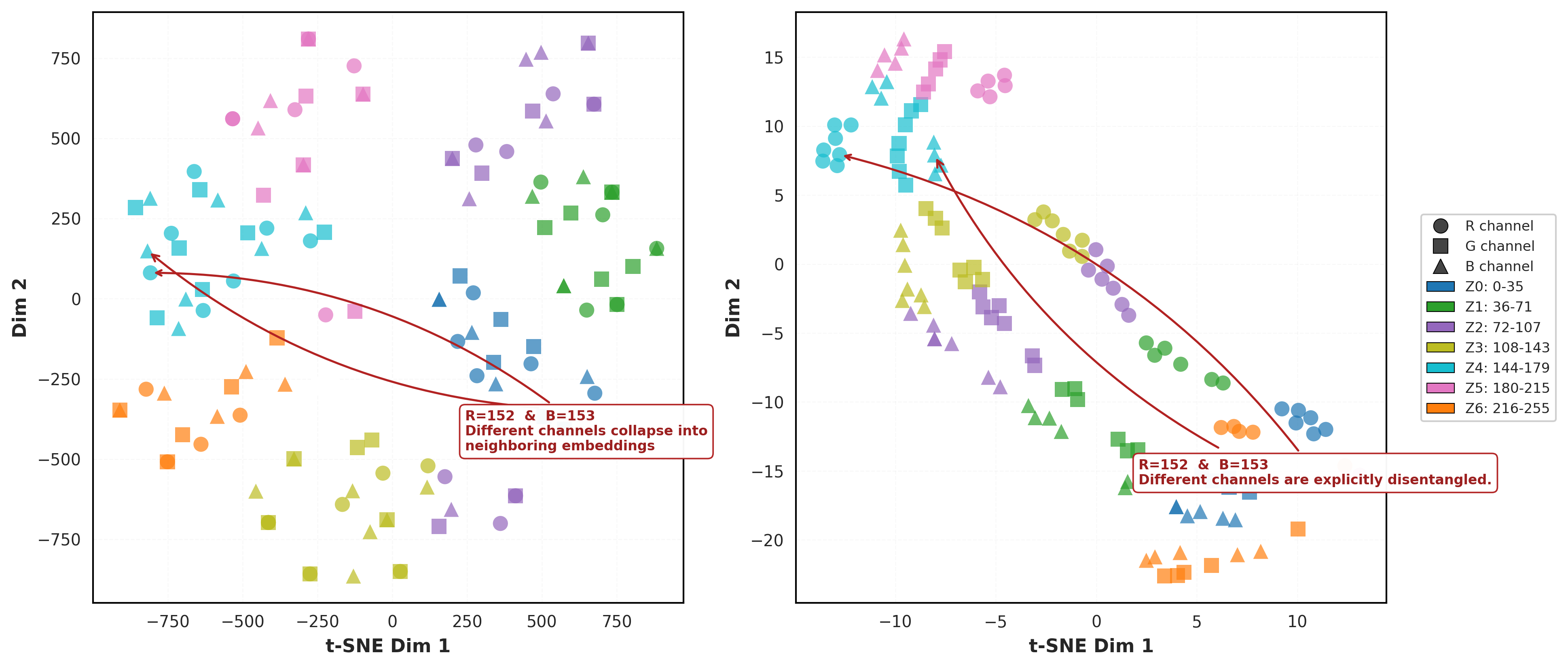}

\caption{Embedding-space visualization of LLaMA tokenizer embeddings (left) and 7-D PixEmb representations (right). We sample 40 RGB pixels in HSV space and visualize their RGB sub-pixels. Marker shapes denote channel identities, while colors denote intensity zones over the scalar range 0--255. In the tokenizer embedding space, numerically related values and channel instances can be scattered or collapsed due to tokenizer-specific numeric representations. In contrast, PixEmb forms more structured neighborhoods by intensity range and channel identity, suggesting better preservation of pixel-level relationships for entropy modeling.}
\label{fig:pixel_embedding_viz}
\end{figure*}
This section evaluates LUMI from four perspectives. We first compare LUMI with classical, learned, and LLM-based lossless codecs under the standard in-domain setting. We then examine unseen-domain generalization, where the target domain is excluded during training and evaluated without test-time adaptation. Finally, we conduct detailed ablations on tokenizer dependence, the proposed components, backbone scale, and training-data efficiency.

\subsection{Experimental Protocol}

\textbf{Datasets.}
Following recent lossless image compression studies, we evaluate RGB images from natural, medical, and remote-sensing domains. Kodak is used as the natural-image benchmark. BRACS\cite{brancati2022bracs} contains histopathology images with fine-grained tissue structures and domain-specific color statistics. BED4RS\cite{BED4RS} contains remote-sensing imagery with large spatial layouts and repetitive geographic patterns. For BED4RS, we use the mountain and forest subsets. Since LUMI is trained and evaluated patch-wise, Table \ref{tab:dataset_split} reports both image counts and patch counts.

\begin{table}[t]
\centering
\caption{Dataset split statistics. Patches are non-overlapping $16\times16$ RGB patches and serve as the effective training and evaluation samples.}
\label{tab:dataset_split}
\begin{tabular}{lccc}
\hline
Dataset & Domain & Images (train/test) & Patches (train/test) \\
\hline
Kodak & Natural & 12/12 & 18,432 / 18,432 \\
BED4RS & Remote sensing & 30/30 & 41,070 / 41,070 \\
BRACS & Medical & 4/8 & 9,767 / 12,655 \\
\hline
\end{tabular}
\end{table}

\textbf{Settings.}
For Kodak, we train on images 1--12 and evaluate on images 13--24 unless otherwise specified. For BRACS, we train on 4 images and evaluate on the rest 8 images. For BED4RS, we use 30 training images and 30 image-disjoint evaluation images from the selected mountain and forest subsets. All train/test partitions are image-disjoint.

\textbf{Training details.}
All input images are converted to RGB and partitioned into non-overlapping $16\times16$ patches. Each patch is flattened in row-major order with channel order $R\rightarrow G\rightarrow B$, yielding a sequence of $16\times16\times3=768$ sub-pixel symbols. Patches smaller than $16\times16$ are discarded. Unless otherwise specified, the default LLM backbones are Llama-3.2-3B \cite{dubey2024llama}, Qwen3-8B \cite{yang2025qwen3}, and Gemma-3-4B \cite{gemmateam2025gemma3}. LUMI trains only a soft prefix of length $P=16$, PixEmb, INP, and the 256-way pixel prediction head, while keeping the LLM backbone frozen. We optimize with AdamW, using learning rates of $5\times10^{-4}$ for PixEmb and the pixel head, $3\times10^{-4}$ for the soft prefix, and $1\times10^{-4}$ for INP. Main experiments are trained for 12 epochs with batch size 4--8 on a single NVIDIA A800 GPU.

\textbf{LLM baseline naming.}
We denote the non-finetuned LLaMA baseline following the \PtwoLLM{} pipeline as LLaMA (\PtwoLLM{} Vanilla). It represents pixel values as textual numeric symbols using the LLaMA tokenizer and prepends the task prompt $q$: \emph{``Every three values denote an RGB color of a single pixel of a flattened two-dimensional image. Predict the next RGB value based on the previous pixels. ''} The prompted sequence is then fed into the frozen LLaMA model, and the pixel distribution is formed by selecting logits corresponding to values 0--255 from the language vocabulary. LLaMA (\PtwoLLM{} LoRA) denotes the same pipeline with LoRA finetuning. We follow the \PtwoLLM{} configuration as closely as possible while using our data split and training schedule for controlled comparison.

\textbf{Metrics.}
We report all compression rates in bits per pixel (BPP), where lower values indicate better compression. All reported rates are obtained from the final lossless compression pipeline, and decoded images are verified to exactly match the original inputs.
\subsection{Tokenizer Fragmentation}
Before evaluating compression performance, we verify whether textual pixel representations are portable across LLM families. We feed the string ``255'' into the native tokenizers of three representative backbones. As shown in Table \ref{tab:tokenization}, LLaMA-3.2 maps the pixel value to a single token, whereas Qwen3 and Gemma-3 fragment the same string into three separate digit tokens. We observe the same tokenization pattern within each major model family across parameter scales, including Qwen3-0.6B, 1.7B, 4B, and 8B.

\begin{table}[t]
\centering
\caption{Tokenization of the string ``255'' across LLM families. Tokenizer-dependent pixel representations are not portable across families.}
\label{tab:tokenization}
\begin{tabular}{lcc}
\hline
Tokenizer & Tokens & Token IDs \\
\hline
Llama-3.2-3B & [``255''] & [3192] \\
Qwen3-0.6B & [``2'',``5'',``5''] & [17,20,20] \\
Gemma-3-4B & [``2'',``5'',``5''] & [236778,236810,236810] \\
\hline
\end{tabular}
\end{table}

This tokenizer test explains why pixel-as-text compression pipelines can become family-specific: the same source symbol may correspond to different sequence lengths and different vocabulary events. LUMI avoids this dependence by bypassing the text tokenizer on the image side and predicting directly over the 256-symbol pixel alphabet.

\subsection{In-Domain Compression Results}
We first evaluate lossless compression performance under the in-domain setting, where training and testing are performed on the same dataset.

\begin{table}[t]
\centering
\caption{In-domain compression performance in BPP. For LUMI, each model is trained and evaluated within the same domain.}
\label{tab:indomain}
\begin{tabular}{lccc}
\hline
Method & Kodak & BRACS & BED4RS \\
\hline
\multicolumn{4}{c}{\textbf{Non-LLM codecs}} \\
\hline
JPEG-XL \cite{alakuijala2019jpeg} & 8.95 & 10.89 & 8.03 \\
DLPR \cite{bai2024deep} & 8.74 & 10.25 & 9.32 \\
\hline
\multicolumn{4}{c}{\textbf{LLM-based codecs}} \\
\hline
LLaMA (\PtwoLLM{} Vanilla) & 12.60 & 15.97 & 17.40 \\
LLaMA (\PtwoLLM{} LoRA) & 8.60 & 10.31 & 8.01 \\
LUMI$_{\mathrm{LLaMA}}$ & \textbf{8.56} & 10.25 & 7.91 \\
LUMI$_{\mathrm{Qwen}}$ & 8.63 & \textbf{10.08} & \textbf{7.88} \\
LUMI$_{\mathrm{Gemma}}$ & 8.70 & 10.15 & 8.00 \\
\hline
\end{tabular}
\end{table}

As shown in Table \ref{tab:indomain}, LUMI achieves competitive in-domain performance across all evaluated domains and all three LLM families. On Kodak, BRACS, and BED4RS, LUMI improves over JPEG-XL. The gains are clear on BRACS, where the best LUMI variant reduces BPP from 10.89 to 10.08, and on BED4RS, where it reduces BPP from 8.03 to 7.88. Averaged over all three datasets and three backbones, LUMI obtains 8.91 BPP, compared with 9.29 for JPEG-XL and 9.44 for DLPR.

The comparison with LLaMA (\PtwoLLM{}) highlights the role of tokenizer-free adaptation. Without LoRA, the \PtwoLLM{}-style pathway is not naturally aligned with RGB pixel distributions. With LoRA, it becomes competitive, but it remains tied to a tokenizer/backbone family whose vocabulary provides suitable numeric tokens. In contrast, LUMI keeps the backbone frozen and matches or improves upon LLaMA (\PtwoLLM{} LoRA) on the evaluated datasets. Similar performance across LLaMA, Qwen, and Gemma suggests that the proposed interface reduces dependence on a specific tokenizer design.

\subsection{Unseen-Domain Generalization Results}

The unseen-domain setting tests whether LUMI can serve as a reusable compression interface when the target image distribution is unavailable during training. For each target domain, we train the lightweight LUMI modules on the other two source domains and evaluate on the held-out target domain without test-time adaptation. For example, the BRACS result is obtained by training on Kodak and BED4RS and then testing on BRACS. All hyperparameters follow the in-domain setting.

\begin{table}[t]
\centering
\caption{Unseen-domain generalization results in BPP. For each target dataset, LUMI is trained on the other two domains and evaluated on the target domain without test-time adaptation.}
\label{tab:unseen}
\begin{tabular}{lccc}
\hline
Method & Kodak & BRACS & BED4RS \\
\hline
\multicolumn{4}{c}{\textbf{Non-LLM codecs}} \\
\hline
JPEG-XL & 8.95 & 10.89 & 8.03 \\
DLPR & 8.74 & 10.25 & 9.32 \\
\hline
\multicolumn{4}{c}{\textbf{LLM-based codecs}} \\
\hline
LLaMA (\PtwoLLM{} Vanilla) & 12.60 & 15.97 & 17.40 \\
LLaMA (\PtwoLLM{} LoRA) & 9.01 & 10.22 & 8.81 \\
LUMI$_{\mathrm{LLaMA}}$ & 8.70 & 10.19 & 7.96 \\
LUMI$_{\mathrm{Qwen}}$ & \textbf{8.68} & \textbf{10.15} & \textbf{7.94} \\
LUMI$_{\mathrm{Gemma}}$ & 8.72 & 10.21 & 7.98 \\
\hline
\end{tabular}
\end{table}

As shown in Table \ref{tab:unseen}, LUMI preserves competitive compression under leave-one-domain-out evaluation. The best Qwen-based model achieves 8.68 BPP on Kodak, 10.15 BPP on BRACS, and 7.94 BPP on BED4RS. The three LUMI variants also produce similar rates across backbones, although their tokenizers and pretraining corpora differ. This supports the claim that once pixel values are represented directly in continuous embedding space and decoded through a 256-way head, compression behavior becomes less dependent on the native text tokenizer.

\subsection{Ablation Studies}

\textbf{Effect of pixel embedding.}
We compare the tokenizer-based numeric interface in \PtwoLLM{} with the tokenizer-free PixEmb--Head interface in LUMI. Since PixEmb bypasses the language tokenizer, it is paired with the 256-way pixel head to produce a native categorical distribution over $\Xset$. We further study feature variants: the 4-D variant removes the RGB channel one-hot vector and only encodes normalized intensity statistics, while the 7-D variant additionally includes explicit RGB channel identity.

\begin{table}[t]
\centering
\caption{Ablation study on PixEmb in BPP. Lower is better.}
\label{tab:pixemb}
\begin{tabular}{lccc}
\hline
Variant & Kodak & BRACS & BED4RS \\
\hline
LLaMA (\PtwoLLM{} Vanilla) & 12.60 & 15.97 & 17.40 \\
w/ 4-D PixEmb & 9.52 & 11.48 & 10.18 \\
w/ 7-D PixEmb & 9.05 & 10.82 & 9.68 \\
\hline
\end{tabular}
\end{table}

As shown in Table~\ref{tab:pixemb}, the tokenizer-free PixEmb--Head interface consistently improves compression across datasets. Compared with LLaMA (\PtwoLLM{} Vanilla), the 7-D variant reduces BPP by 3.55 on Kodak, 5.15 on BRACS, and 7.72 on BED4RS. This indicates that a structured pixel-native representation and a native 256-way prediction head are more suitable for entropy modeling than textual numeric tokenization. The improvement from the 4-D to the 7-D variant further shows the value of explicit channel identity.

The visualization in Figure~\ref{fig:pixel_embedding_viz} further supports this finding. Compared with numeric tokenizer embeddings, PixEmb yields a more structured embedding space, where sub-pixel symbols are organized according to intensity ranges and RGB channel identities rather than arbitrary tokenizer segmentation.

\begin{figure*}[htbp]

\centering

\includegraphics[width=\textwidth]{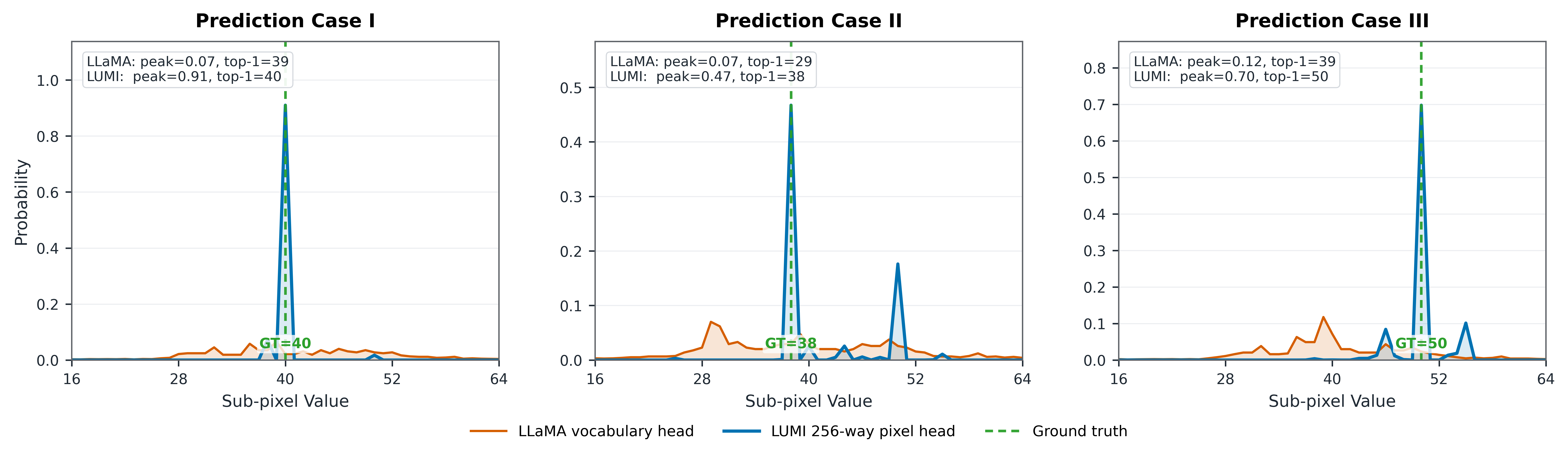}

\caption{Representative sub-pixel prediction distributions on BED4RS. The three panels show consecutive sub-pixel predictions corresponding to one complete RGB pixel. We compare LLaMA (\PtwoLLM{} Vanilla), which obtains pixel probabilities by selecting logits of tokens 0--255 from the LLaMA vocabulary head and normalizing them over the numeric-token subset, with a LUMI variant using only PixEmb and the 256-way pixel head. The green dashed line marks the ground-truth sub-pixel value. The full-vocabulary numeric-token mass of LLaMA is 72.61\%, 86.44\%, and 73.15\% for Cases I--III, respectively, yet the resulting selected-token distributions remain less aligned with the ground truth than the native 256-way pixel-head distributions.}
\label{fig:pred_dist}

\end{figure*}

\textbf{Effect of pixel head.}
To qualitatively examine the output side of the tokenizer-free interface, Figure~\ref{fig:pred_dist} shows three consecutive sub-pixel predictions from BED4RS, corresponding to one complete RGB pixel. We compare LLaMA (\PtwoLLM{} Vanilla), which follows the textual numeric interface and derives pixel probabilities by selecting the logits of tokens 0--255 from the LLaMA vocabulary head, with a LUMI variant using only PixEmb and the 256-way pixel head. Across the three cases, the PixEmb--Head variant places sharp probability peaks at the ground-truth sub-pixel values, whereas the vocabulary-head distribution remains diffuse and yields misaligned top-1 predictions. Although LLaMA assigns a substantial fraction of full-vocabulary probability mass to numeric tokens, with an average numeric mass of 77.40\% across the three cases, the selected-token distribution is still suboptimally aligned with the ground-truth pixel values. This suggests that a native 256-symbol pixel head provides a more appropriate output interface for entropy coding.

\textbf{Effect of intra-patch position encoding.}
We next evaluate INP, which introduces explicit two-dimensional coordinates into the pixel representation. Unlike PixEmb, which models intensity and channel structure, INP provides a spatial inductive bias to compensate for the flattening of the 2-D patch into a 1-D sequence.

\begin{table}[t]
\centering
\caption{Ablation study on INP in BPP.}
\label{tab:inp}
\begin{tabular}{lccc}
\hline
Variant & Kodak & BRACS & BED4RS \\
\hline
LLaMA (\PtwoLLM{} Vanilla) & 12.60 & 15.97 & 17.40 \\
w/ INP & 9.48 & 11.45 & 10.12 \\
\hline
\end{tabular}
\end{table}

Table \ref{tab:inp} shows that INP improves compression on all datasets, reducing BPP by 3.12 on Kodak, 4.52 on BRACS, and 7.28 on BED4RS relative to the tokenizer-based baseline. These gains indicate that explicit spatial coordinates help the frozen LLM exploit local image dependencies that are otherwise implicit in the flattened sequence.

\textbf{Task Prompt versus Soft Prefix.}
We further study LUMI from the perspective of LLM reprogramming \cite{ye2025reprogrammability}, where a frozen language model is adapted to a new task via input-space transformations rather than parameter updates. In this view, the task prompt (TP) and soft prefix (SP) act as two complementary reprogramming mechanisms: TP provides a discrete symbolic instruction in natural language, while SP serves as a continuous learnable transformation in the embedding space.

We evaluate their interaction under the default tokenizer-based LLaMA setting on the Kodak dataset. Specifically, the model is trained on the Kodak training split for 12 epochs using LLaMA-3.2-3B as the frozen backbone and evaluated on the held-out test split. In this ablation, only the task prompt and soft prefix are involved during training and evaluation; PixEmb, INP, and the 256-way prediction head are not used. When SP is enabled, only the soft-prefix parameters are optimized, while the LLM backbone remains fixed.

\begin{table}[t]
\centering
\caption{Textual task-prompt (TP) and soft-prefix (SP) ablation under the LLM reprogramming perspective on Kodak. Only SP is optimized when enabled, and no PixEmb, INP, or 256-way head is trained.}
\label{tab:prompt_prefix}
\begin{tabular}{lccc}
\hline
SP setting & TP in Train  & TP in Test & BPP \\
\hline
None & -- & \xmark & 17.73 \\
None & -- & \cmark & 12.61 \\
Trained & \xmark & \xmark & 14.29 \\
Trained & \xmark & \cmark & 11.61 \\
Trained & \cmark & \xmark & 9.82 \\
Trained & \cmark & \cmark & \textbf{9.43} \\
% None & -- & No & 17.73 \\
% None & -- & Yes & 12.61 \\
% Trained & No & No & 14.29 \\
% Trained & No & Yes & 11.61 \\
% Trained & Yes & No & 9.82 \\
% Trained & Yes & Yes & \textbf{9.43} \\
\hline
\end{tabular}
\end{table}

As shown in Table~\ref{tab:prompt_prefix}, both TP and SP independently contribute to compression performance improvements under the reprogramming framework. The task prompt alone provides a strong gain even without training, indicating that explicit textual instructions induce a useful prior for pixel-level autoregressive modeling. The soft prefix further improves performance, demonstrating that learned continuous reprogramming in the embedding space captures additional structure beyond discrete prompting.

More importantly, combining TP and SP leads to consistent improvements, suggesting that symbolic and continuous reprogramming operate on complementary aspects of the input space. Finally, training the soft prefix with task-prompt conditioning yields the best performance, indicating that alignment between discrete (TP) and continuous (SP) reprogramming signals is critical for optimal entropy modeling in frozen LLMs.

Overall, these results support the interpretation of LUMI as an input-space reprogramming framework, where lossless image compression emerges from structured transformation of pixel sequences rather than modification of model parameters.

\textbf{Cross-domain transferability of PixEmb and INP.}
Finally, we evaluate cross-domain transferability. PixEmb and INP are trained on Kodak with Llama-3.2-3B, and directly applied to BRACS and BED4RS during inference without finetuning.

\begin{table}[t]
\centering
\caption{Cross-domain transfer of Kodak-trained PixEmb and INP in BPP.}
\label{tab:cross_domain_components}
\begin{tabular}{lcc}
\hline
Variant & BRACS & BED4RS \\
\hline
LLaMA (\PtwoLLM{} Vanilla) & 15.97 & 17.40 \\
w/ Kodak PixEmb & 11.95 & 10.55 \\
w/ Kodak PixEmb + INP & \textbf{11.72} & \textbf{10.38} \\
\hline
\end{tabular}
\end{table}

Table \ref{tab:cross_domain_components} shows that PixEmb significantly reduces BPP on both target domains, indicating that it learns a reusable pixel-to-latent interface rather than dataset-specific statistics. INP further improves performance consistently, showing that spatial inductive biases also transfer across domains.

Overall, PixEmb and INP form a domain-agnostic pixel representation layer that generalizes across heterogeneous image distributions within a frozen LLM framework.

\subsection{Model and Data Scalability}

One motivation for LUMI is to examine whether a tokenizer-free interface can inherit some scalability from foundation models. We study scalability from two perspectives: increasing frozen backbone capacity and increasing adaptation data.

\textbf{Model Scaling}
We conduct this study on BRACS using the Qwen3 family as frozen backbones. The training protocol is kept identical across all model sizes to isolate the effect of backbone capacity. Specifically, all models are trained on the BRACS training split and evaluated on the fixed test split for 12 epochs. The trainable LUMI components, including the pixel embedding module, intra-patch position encoding, soft prefix, and 256-way prediction head, remain identical for all backbone scales.

\begin{table}[htbp]
\centering
\small
\caption{LLM scaling ablation on BRACS.}
\label{tab:llm_scaling}
\setlength{\tabcolsep}{4pt}
\begin{tabular}{l c}
\hline
\textbf{Method} & \textbf{BPP$\downarrow$} \\
\hline
JPEG-XL & 10.89 \\
DLPR & 10.25 \\
\hline
LUMI (Qwen3-0.6B) & 11.58 \\
LUMI (Qwen3-1.7B) & 11.26 \\
LUMI (Qwen3-4B) & 11.03 \\
LUMI (Qwen3-8B) & 10.08 \\
LUMI (Qwen3-14B) & \textbf{9.98} \\
\hline
\end{tabular}
\end{table}

As shown in Table~\ref{tab:llm_scaling}, LUMI exhibits consistent performance improvements as the frozen backbone scales from 0.6B to 14B parameters. Even the smallest 0.6B model substantially outperforms the tokenizer-dependent LLaMA (\PtwoLLM{} Vanilla) baseline on this domain, reducing the compression rate from 15.97 to 11.58 BPP. Increasing the backbone to 8B further reduces the rate to 10.08 BPP, already surpassing DLPR under the same evaluation setting, while the 14B model achieves a further improvement to 9.98 BPP. Although the improvement from 8B to 14B becomes smaller than that from 4B to 8B, the overall trend remains consistently positive.

These observations indicate that our proposed tokenizer-free pixel interface is not tied to a particular model scale, but can effectively leverage increasingly capable frozen foundation models without modifying the compression framework itself. 

\textbf{Data scaling.}
We next vary the number of BRACS training patches while using Qwen3-8B as the frozen backbone. Training patches are sampled from the BRACS training split, and evaluation is performed on the fixed BRACS test split. We vary the training set from 2,000 patches to the full training split of 9,767 patches, and train all models for 12 epochs under the same optimization setting.

\begin{table}[htbp]
\centering
\small
\caption{Data scaling analysis on BRACS}
\label{tab:data_efficient}
\setlength{\tabcolsep}{4pt}
\begin{tabular}{l c c}
\hline
\textbf{Method} & \textbf{Training patches} & \textbf{BPP$\downarrow$} \\
\hline
JPEG-XL   & - & 10.89 \\
DLPR      & - & 10.25 \\
LLaMA(P2LLM Vanilla)      & - & 15.97 \\
\hline
LUMI (Qwen3-8B) & 2000 patches  & 11.84 \\
LUMI (Qwen3-8B) & 4000 patches  & 11.39 \\
LUMI (Qwen3-8B) & 6000 patches  & 10.94 \\
LUMI (Qwen3-8B) & 8000 patches  & 10.67 \\
LUMI (Qwen3-8B) & 9767 patches& \textbf{10.08} \\
\hline
\end{tabular}
\end{table}

As shown in Table~\ref{tab:data_efficient}, the compression performance of LUMI improves consistently as more adaptation data become available. Using only 2,000 training patches, LUMI achieves 11.84 BPP, already substantially outperforming the tokenizer-dependent LlaMA baseline on this domain. Increasing the training set to 6,000 patches further reduces the compression rate to 10.94 BPP, bringing the performance close to JPEG-XL. With the full training split (9,767 patches), LUMI reaches 10.08 BPP, slightly surpassing DLPR while adapting only lightweight pixel-space modules on top of a frozen Qwen3-8B backbone.

Together with the model-scaling study, these results demonstrate that LUMI scales consistently with both model capacity and adaptation data, showing that LUMI provides a scalable interface for deploying heterogeneous frozen foundation models as universal entropy models for lossless image compression.

\subsection{Limitations}

Although LUMI achieves competitive compression performance, several limitations remain.
The current framework models patches independently and therefore cannot exploit inter-patch dependencies.
Autoregressive LLM inference also incurs substantially higher decoding latency than conventional codecs \cite{deletang2024language,chen2026large}.
Finally, since the LLM backbone is kept frozen, the proposed interface improves only how image symbols are presented to the model, while the achievable compression performance remains fundamentally limited by the backbone's intrinsic predictive capability.
Future work will investigate hierarchical context modeling, faster decoding, and more efficient context reuse across neighboring patches.

\section{Conclusions}

This paper presented LUMI, a model-agnostic framework for lossless image compression with frozen LLM backbones. LUMI maps pixel intensity, channel identity, and intra-patch position directly into the LLM embedding space, and predicts a 256-way distribution for arithmetic coding without modifying the backbone. Experiments on natural, medical, and remote-sensing images show competitive compression rates and consistent behavior across LLaMA, Qwen, and Gemma. These results suggest that frozen foundation models can serve as reusable entropy models when equipped with an image-symbol interface for lossless image coding.

\bibliographystyle{IEEEtran}
\bibliography{main}

@inproceedings{chen2026large,
  title     = {Large Language Models for Lossless Image Compression: Next-Pixel Prediction in Language Space is All You Need},
  author    = {Chen, Kecheng and Zhang, Pingping and Liu, Hui and Liu, Jie and Liu, Yibing and Huang, Jiaxin and Wang, Shiqi and Yan, Hong and Li, Haoliang},
  booktitle = {The Thirty-ninth Annual Conference on Neural Information Processing Systems},
  year      = {2025},
  url       = {https://openreview.net/forum?id=FXBBy1caOX}
}

@article{li2025lossless,
  title     = {Lossless data compression by large models},
  author    = {Li, Ziguang and Huang, Chao and Wang, Xuliang and Hu, Haibo and Wyeth, Cole and Bu, Dongbo and Yu, Quan and Gao, Wen and Liu, Xingwu and Li, Ming},
  journal   = {Nature Machine Intelligence},
  volume    = {7},
  number    = {5},
  pages     = {794--799},
  year      = {2025},
  doi       = {10.1038/s42256-025-01033-7},
  url       = {https://doi.org/10.1038/s42256-025-01033-7},
  publisher = {Nature Publishing Group}
}

@misc{BED4RS,
  title        = {BED4RS: Benchmark Datasets for Remote Sensing Image Understanding},
  author       = {{CAPTAIN-WHU}},
  year         = {2020},
  howpublished = {\url{https://captain-whu.github.io/BED4RS/}},
  note         = {Accessed: 2026-07-05}
}

@article{du2025visualprompts,
  title   = {Large Language Model for Lossless Image Compression with Visual Prompts},
  author  = {Du, Junhao and Zhou, Chuqin and Cao, Ning and Chen, Gang and Chen, Yunuo and Cheng, Zhengxue and Song, Li and Lu, Guo and Zhang, Wenjun},
  journal = {arXiv preprint arXiv:2502.16163},
  year    = {2025},
  url     = {https://arxiv.org/abs/2502.16163},
  doi     = {10.48550/arXiv.2502.16163}
}

@inproceedings{deletang2024language,
  title     = {Language Modeling Is Compression},
  author    = {Del{\'e}tang, Gr{\'e}goire and Ruoss, Anian and Duquenne, Paul-Ambroise and Catt, Elliot and Genewein, Tim and Mattern, Christopher and Grau-Moya, Jordi and Wenliang, Li Kevin and Aitchison, Matthew and Orseau, Laurent and Hutter, Marcus and Veness, Joel},
  booktitle = {The Twelfth International Conference on Learning Representations},
  year      = {2024},
  url       = {https://openreview.net/forum?id=jznbgiynus}
}

@inproceedings{alakuijala2019jpeg,
  author = {Jyrki Alakuijala and Ruud Van Asseldonk and Sami Boukortt and Martin Bruse and Iulia-Maria Com{\c{s}}a and Moritz Firsching and Thomas Fischbacher and Evgenii Kliuchnikov and Sebastian Gomez and Robert Obryk and others},
  title = {Jpeg xl next-generation image compression architecture and coding tools},
  booktitle = {Applications of digital image processing XLII},
  pages = {112--124},
  publisher = {SPIE},
  year = {2019}
}

@article{ali2023tokenizer,
  author = {Mehdi Ali and Michael Fromm and Klaudia Thellmann and Richard Rutmann and Max L{\"u}bbering and Johannes Leveling and Katrin Klug and Jan Ebert and Niclas Doll and Jasper Schulze Buschhoff and others},
  title = {Tokenizer Choice For {LLM} Training: Negligible or Crucial?},
  journal = {arXiv preprint arXiv:2310.08754},
  year = {2024},
  url = {https://arxiv.org/abs/2310.08754},
  doi = {10.48550/arXiv.2310.08754}
}

@article{bai2024deep,
  author = {Yuanchao Bai and Xianming Liu and Kai Wang and Xiangyang Ji and Xiaolin Wu and Wen Gao},
  title = {Deep lossy plus residual coding for lossless and near-lossless image compression},
  journal = {IEEE Transactions on Pattern Analysis and Machine Intelligence},
  volume = {46},
  number = {5},
  pages = {3577--3594},
  year = {2024},
  doi = {10.1109/TPAMI.2023.3348486}
}

@techreport{boutell1997png,
  author = {Thomas Boutell},
  title = {Png (portable network graphics) specification version 1.0},
  institution = {World Wide Web Consortium},
  year = {1997}
}

@article{dubey2024llama,
  author = {{Llama Team, AI at Meta}},
  title = {The {Llama} 3 Herd of Models},
  journal = {arXiv preprint arXiv:2407.21783},
  year = {2024},
  url = {https://arxiv.org/abs/2407.21783},
  doi = {10.48550/arXiv.2407.21783}
}

@article{yang2025qwen3,
  author = {Yang, An and Li, Anfeng and Yang, Baosong and Zhang, Beichen and Hui, Binyuan and Zheng, Bo and Yu, Bowen and Gao, Chang and Huang, Chengen and Lv, Chenxu and others},
  title = {{Qwen3} Technical Report},
  journal = {arXiv preprint arXiv:2505.09388},
  year = {2025},
  url = {https://arxiv.org/abs/2505.09388},
  doi = {10.48550/arXiv.2505.09388}
}

@article{gemmateam2025gemma3,
  author = {{Gemma Team}},
  title = {{Gemma} 3 Technical Report},
  journal = {arXiv preprint arXiv:2503.19786},
  year = {2025},
  url = {https://arxiv.org/abs/2503.19786},
  doi = {10.48550/arXiv.2503.19786}
}

@article{heurtel2024compression,
  author = {David Heurtel-Depeiges and Anian Ruoss and Joel Veness and Tim Genewein},
  title = {Compression via pre-trained transformers: A study on byte-level multimodal data},
  journal = {arXiv preprint arXiv:2410.05078},
  year = {2024}
}

@inproceedings{hu2021lora,
  author = {Edward J Hu and Yelong Shen and Phillip Wallis and Zeyuan Allen-Zhu and Yuanzhi Li and Shean Wang and Lu Wang and Weizhu Chen},
  title = {{LoRA}: Low-Rank Adaptation of Large Language Models},
  booktitle = {International Conference on Learning Representations},
  year = {2022},
  url = {https://openreview.net/forum?id=nZeVKeeFYf9}
}

@inproceedings{li2021prefix,
  author = {Li, Xiang Lisa and Liang, Percy},
  title = {Prefix-Tuning: Optimizing Continuous Prompts for Generation},
  booktitle = {Proceedings of the 59th Annual Meeting of the Association for Computational Linguistics and the 11th International Joint Conference on Natural Language Processing},
  pages = {4582--4597},
  year = {2021},
  doi = {10.18653/v1/2021.acl-long.353},
  url = {https://aclanthology.org/2021.acl-long.353}
}

@inproceedings{lester2021power,
  author = {Lester, Brian and Al-Rfou, Rami and Constant, Noah},
  title = {The Power of Scale for Parameter-Efficient Prompt Tuning},
  booktitle = {Proceedings of the 2021 Conference on Empirical Methods in Natural Language Processing},
  pages = {3045--3059},
  year = {2021},
  doi = {10.18653/v1/2021.emnlp-main.243},
  url = {https://aclanthology.org/2021.emnlp-main.243}
}

@inproceedings{mentzer2019pract,
  author = {Fabian Mentzer and Eirikur Agustsson and Michael Tschannen and Radu Timofte and Luc Van Gool},
  title = {Practical full resolution learned lossless image compression},
  booktitle = {Proceedings of the IEEE/CVF conference on computer vision and pattern recognition},
  pages = {10629--10638},
  year = {2019}
}

@inproceedings{mentzer2020learning,
  author = {Fabian Mentzer and Luc Van Gool and Michael Tschannen},
  title = {Learning better lossless compression using lossy compression},
  booktitle = {Proceedings of the IEEE/CVF conference on computer vision and pattern recognition},
  pages = {6638--6647},
  year = {2020}
}

@article{salimans2017pixelcnn,
  author = {Tim Salimans and Andrej Karpathy and Xi Chen and Diederik P Kingma},
  title = {Pixelcnn++: Improving the pixelcnn with discretized logistic mixture likelihood and other modifications},
  journal = {arXiv preprint arXiv:1701.05517},
  year = {2017}
}

@article{brancati2022bracs,
  author  = {Nadia Brancati and Anna Maria Anniciello and Pushpak Pati and
             Daniel Riccio and Giosu{\`e} Scognamiglio and Guillaume Jaume and
             Giuseppe De Pietro and Maurizio Di Bonito and
             Antonio Foncubierta-Rodr{\'i}guez and Gerardo Botti and
             Maria Gabrani and Florinda Feroce and Maria Frucci},
  title   = {BRACS: A Dataset for BReAst Carcinoma Subtyping in H\&E Histology Images},
  journal = {Database},
  volume  = {2022},
  pages   = {baac093},
  year    = {2022},
  doi     = {10.1093/database/baac093}
}

@inproceedings{van2016pixel,
  author = {A{\"a}ron Van Den Oord and Nal Kalchbrenner and Koray Kavukcuoglu},
  title = {Pixel recurrent neural networks},
  booktitle = {International conference on machine learning},
  pages = {1747--1756},
  publisher = {PMLR},
  year = {2016}
}

@article{weinberger2000loco,
  author = {Marcelo J Weinberger and Gadiel Seroussi and Guillermo Sapiro},
  title = {The loco-i lossless image compression algorithm: Principles and standardization into jpeg-ls},
  journal = {IEEE Transactions on Image processing},
  volume = {9},
  number = {8},
  pages = {1309--1324},
  year = {2000}
}

@article{zhang2021iflow,
  author = {Shifeng Zhang and Ning Kang and Tom Ryder and Zhenguo Li},
  title = {iflow: Numerically invertible flows for efficient lossless compression via a uniform coder},
  journal = {Advances in Neural Information Processing Systems},
  volume = {34},
  pages = {5822--5833},
  year = {2021}
}

@article{ye2025reprogrammability,
  title={Neural Network Reprogrammability: A Unified Theme on Model Reprogramming, Prompt Tuning, and Prompt Instruction}, 
      author={Zesheng Ye and Chengyi Cai and Ruijiang Dong and Jianzhong Qi and Lei Feng and Pin-Yu Chen and Feng Liu},
      year={2025},
      eprint={2506.04650},
      archivePrefix={arXiv},
      primaryClass={cs.LG},
      url={https://arxiv.org/abs/2506.04650}, 
}

@inproceedings{tancik2020fourier,
  title     = {Fourier Features Let Networks Learn High Frequency Functions in Low Dimensional Domains},
  author    = {Tancik, Matthew and Srinivasan, Pratul P. and Mildenhall, Ben and Fridovich-Keil, Sara and Raghavan, Nithin and Singhal, Utkarsh and Ramamoorthi, Ravi and Barron, Jonathan T. and Ng, Ren},
  booktitle = {Advances in Neural Information Processing Systems (NeurIPS)},
  year      = {2020},
  url       = {https://arxiv.org/abs/2006.10739}
}

@inproceedings{zhang2021ivpf,
  author = {Shifeng Zhang and Chen Zhang and Ning Kang and Zhenguo Li},
  title = {ivpf: Numerical invertible volume preserving flow for efficient lossless compression},
  booktitle = {Proceedings of the IEEE/CVF Conference on Computer Vision and Pattern Recognition},
  pages = {620--629},
  year = {2021}
}

@inproceedings{zhang2024learned,
  author = {Zhe Zhang and Huairui Wang and Zhenzhong Chen and Shan Liu},
  title = {Learned lossless image compression based on bit plane slicing},
  booktitle = {Proceedings of the IEEE/CVF Conference on Computer Vision and Pattern Recognition},
  pages = {27579--27588},
  year = {2024}
}

@article{chen2021evaluating,
  title   = {Evaluating Large Language Models Trained on Code},
  author  = {Chen, Mark and Tworek, Jerry and Jun, Heewoo and Yuan, Qiming and Pinto, Henrique Ponde de Oliveira and Kaplan, Jared and Edwards, Harri and Burda, Yuri and Joseph, Nicholas and Brockman, Greg and others},
  journal = {arXiv preprint arXiv:2107.03374},
  year    = {2021}
}

@inproceedings{wei2022chain,
  title     = {Chain-of-Thought Prompting Elicits Reasoning in Large Language Models},
  author    = {Wei, Jason and Wang, Xuezhi and Schuurmans, Dale and Bosma, Maarten and Ichter, Brian and Xia, Fei and Chi, Ed H. and Le, Quoc V. and Zhou, Denny},
  booktitle = {Advances in Neural Information Processing Systems},
  volume    = {35},
  pages     = {24824--24837},
  year      = {2022}
}

@inproceedings{lewis2020retrieval,
  title     = {Retrieval-Augmented Generation for Knowledge-Intensive NLP Tasks},
  author    = {Lewis, Patrick and Perez, Ethan and Piktus, Aleksandra and Petroni, Fabio and Karpukhin, Vladimir and Goyal, Naman and K{\"u}ttler, Heinrich and Lewis, Mike and Yih, Wen-tau and Rockt{\"a}schel, Tim and others},
  booktitle = {Advances in Neural Information Processing Systems},
  volume    = {33},
  pages     = {9459--9474},
  year      = {2020}
}

@inproceedings{alayrac2022flamingo,
  title     = {Flamingo: A Visual Language Model for Few-Shot Learning},
  author    = {Alayrac, Jean-Baptiste and Donahue, Jeff and Luc, Pauline and Miech, Antoine and Barr, Iain and Hasson, Yana and Lenc, Karel and Mensch, Arthur and Millican, Katherine and Reynolds, Malcolm and others},
  booktitle = {Advances in Neural Information Processing Systems},
  volume    = {35},
  pages     = {23716--23736},
  year      = {2022}
}

@book{cover2006elements,
  title     = {Elements of Information Theory},
  author    = {Cover, Thomas M. and Thomas, Joy A.},
  edition   = {2},
  publisher = {Wiley-Interscience},
  year      = {2006}
}

@article{witten1987arithmetic,
  title   = {Arithmetic Coding for Data Compression},
  author  = {Witten, Ian H. and Neal, Radford M. and Cleary, John G.},
  journal = {Communications of the ACM},
  volume  = {30},
  number  = {6},
  pages   = {520--540},
  year    = {1987}
}

\end{document}